\pdfoutput=1

\documentclass[11pt]{article}

\usepackage{EMNLP2023}
\usepackage{times}
\usepackage{graphicx}
\usepackage{latexsym}
\usepackage{subcaption}
\usepackage{tikz}
\usepackage{xspace}
\usepackage{tabularray}
\usepackage{booktabs}
\usepackage{lscape}
\usepackage[T1]{fontenc}

\usepackage[utf8]{inputenc}

\usepackage{microtype}
\usepackage{adjustbox}
\usepackage{inconsolata}
\usepackage{booktabs}
\usepackage{amsmath} 
\usepackage{xspace}
\usepackage[symbol*]{footmisc}
\newcommand{\ignore}[1]{}
\newcommand{\ignorebecauseofproduct}[1]{}
\newcommand{\predicate}[1]{{\tt #1}}
\newcommand{\triple}[1]{{\tt #1}}

\newcommand{\system}{ODKE\,+\xspace}

\title{ODKE+: Ontology-Guided Open-Domain Knowledge Extraction with LLMs}

\author{
Samira Khorshidi\thanks{\texttt{samiraa@apple.com}},\quad
Azadeh Nikfarjam\thanks{\texttt{anikfarjam@apple.com}},\quad
Suprita Shankar\thanks{\texttt{suprita.shankar@apple.com}},\quad
Yisi Sang\thanks{\texttt{yisi\_sang@apple.com}},\\
Yash Govind\thanks{\texttt{yash\_govind@apple.com}},\quad
Hyun Jang\thanks{\texttt{kwanghyun\_jang@apple.com}},\quad
Ali Kasgari\thanks{\texttt{a\_kasgari@apple.com}}, \quad
Alexis McClimans\thanks{\texttt{ak\_mcclimans@apple.com}},\\
Mohamed Soliman\thanks{\texttt{mohamed\_soliman@apple.com}},\quad
Vishnu Konda\thanks{\texttt{vkonda@apple.com}},\quad
Ahmed Fakhry\thanks{\texttt{afakhry@apple.com}},\quad
Xiaoguang Qi\thanks{\texttt{xiaoguang\_qi@apple.com}} \\
Apple Inc.
}

\begin{document}
\maketitle

\begin{abstract}

Knowledge graphs (KGs) are foundational to many AI applications, but maintaining their freshness and completeness remains costly. We present \textbf{ODKE+}, a production-grade system that automatically extracts and ingests millions of open-domain facts from web sources with high precision.

ODKE+ combines modular components into a scalable pipeline: (1) the \textit{Extraction Initiator} detects missing or stale facts, (2) the \textit{Evidence Retriever} collects supporting documents, (3) hybrid \textit{Knowledge Extractors} apply both pattern-based rules and ontology-guided prompting for large language models (LLMs), (4) a lightweight \textit{Grounder} validates extracted facts using a second LLM, and (5) the \textit{Corroborator} ranks and normalizes candidate facts for ingestion.

ODKE+ dynamically generates ontology snippets tailored to each entity type to align extractions with schema constraints, enabling scalable, type-consistent fact extraction across 195 predicates. The system supports batch and streaming modes, processing over 9 million Wikipedia pages and ingesting 19 million high-confidence facts with 98.8\% precision. 

ODKE+ significantly improves coverage over traditional methods, achieving up to 48\% overlap with third-party KGs and reducing update lag by 50 days on average. Our deployment demonstrates that LLM-based extraction, grounded in ontological structure and verification workflows, can deliver trustworthy, production-scale knowledge ingestion with broad real-world applicability.

A recording of the system demonstration is included with the submission and is also available at {\color{blue}\url{https://youtu.be/UcnE3_GsTWs}}.
{\newline}
\end{abstract}

\section{Introduction}

 A knowledge graph (KG) organizes open-domain information into structured representations by capturing semantic relationships among entities. Ontologies provide the backbone of a knowledge graph by defining entity types, properties, and relationships, establishing the semantic framework that governs how knowledge is stored and interpreted. Built on this foundation, knowledge graphs enable a wide range of real-world applications, such as question answering, entity disambiguation, and relationship extraction. The quality and completeness of a KG directly influence the performance of such downstream applications.

Maintaining a KG that is both comprehensive and up-to-date is an ongoing challenge. Traditionally, accurate knowledge ingestion has relied on labor-intensive and costly human curation, which is not scalable. To address this, an automated framework is therefore needed, one that can continuously update KGs with precise facts while ensuring correctness and consistency.

However, building such a system involves overcoming several key challenges:
\begin{itemize}

\item {\textbf{Volume:}} The web contains billions of pages and millions of Wikipedia articles; ingestion must scale accordingly.
\item {\textbf{Variety:}} Information appears in diverse forms, from plain text to semi-structured infoboxes and web tables, requiring flexible extraction techniques.
\item {\textbf{Veracity:}} The web is noisy, with conflicting or outdated information; ensuring accuracy requires evidence selection and trust assessment.
\item {\textbf{Velocity:}} Facts change frequently; keeping KGs fresh necessitates both periodic and near-real-time updates for high-interest topics. 
\end{itemize}
 
\section{Background}
While each of these challenges, volume, variety, veracity, and velocity, has been tackled individually in previous research, few systems offer a unified, production-grade solution for open-domain KG maintenance. Traditional methods for extracting facts from semi-structured sources, such as Wikipedia infoboxes, rely heavily on rule-based extractors that are difficult to scale and generalize.

Prior work on LLM-based knowledge graph construction exhibits three key limitations. First, most systems operate in a batch-only fashion, processing a static corpus without accounting for the frequent and continuous changes that occur on the web \cite{desantis2025testData, ZeroFew2024, KGGen2025, EDC2024, OneKe2025}. As a result, the constructed knowledge graphs quickly become stale and may miss new or updated facts. Second, several approaches lack corroboration mechanisms to verify whether the extracted triples are grounded in source evidence, leaving them vulnerable to hallucinations commonly associated with large language models \cite{iText2KG2024, KGGen2025}. Finally, many systems do not incorporate ontology-guided prompting, which is essential to maintain semantic consistency between the extracted facts and the ontology schema \cite{desantis2025testData, KGGen2025, ZeroFew2024}.

A notable prior effort is the search-based KG completion system by \citep{kb-completion-search-based}, which leverages search query logs to identify knowledge gaps. However, such systems typically lack broad coverage and do not incorporate modern LLM-based extraction techniques or semantic alignment with formal ontologies.

To address these limitations, we present \system(Open Domain Knowledge Extraction), a scalable system that automatically extracts facts from open-domain sources using large language models. It supports batch and streaming modes, employs ontology-guided prompting for semantic consistency, and includes modules for evidence retrieval, hybrid extraction, corroboration, normalization, and ingestion into large-scale knowledge graphs.
This demo showcases the \system pipeline in action; identifying outdated or missing facts, retrieving evidence, extracting ontology-aligned facts, and updating the knowledge graph in near real time. We also share key design choices, operational lessons, and deployment insights from real-world use.
\subsection{Comparison with Prior Versions}
\label{sec:comparison-v123}

The initial version of \system was introduced in \cite{10.1145/3555041.3589672} as the extraction component of the Saga system \cite{saga_sigmod}. ODKE v2 \cite{odke_2023} extended the system with multilingual support, streaming ingestion, and limited LLM-based extractors.

In this paper, we present \textbf{ODKE v3 (\system)}, which includes production-grade components such as grounding verification using a lightweight LLM, dynamic ontology-guided prompting, and expanded predicate support. Table~\ref{tab:odke-versions} summarizes the key evolution across versions. 
\begin{table}[htb]
\centering
\begin{adjustbox}{width=\linewidth}
\begin{tabular}{|l|c|c|c|}
\hline
\textbf{Capability} & \textbf{ODKE v1} & \textbf{ODKE v2} & \textbf{\system(This work)} \\
\hline
Evidence Retrieval & Search-based & Search + Crawl & Crawl \\
Extraction Power & Pattern-based & Pattern + LLM & Pattern + Ontology-guided LLM \\
Multilingual Support & No & Yes & Yes + Locale linking \\
Link Inference & No & Yes & ML-based linking \\
Streaming Support & No & Yes & Yes\\
Stability & Up to 5k/min & Up to 100k/min & 100k+/min \\
Ontology Prompting & -- & Static & Dynamic \\
Grounding Verification & -- & -- & LLM verifier \\
Predicate Coverage & <50 & <50 & 195+ \\
\hline
\end{tabular}
\end{adjustbox}
\caption{Comparison of ODKE v1, ODKE v2, and \system.}
\label{tab:odke-versions}
\end{table}
\section{System Architecture}\label{sec:arch}
\begin{figure}[htb] 
  \begin{subfigure}{\columnwidth}
    \centering
    \resizebox{\columnwidth}{!}{
        \input{Figures/architecture}
    } 
     \caption{ODKE system Overview}
    \label{fig:architecture}
    \vspace{-2mm}
  \end{subfigure}%
   
 \begin{subfigure}{\columnwidth}
    \centering
  \includegraphics[width=\columnwidth]{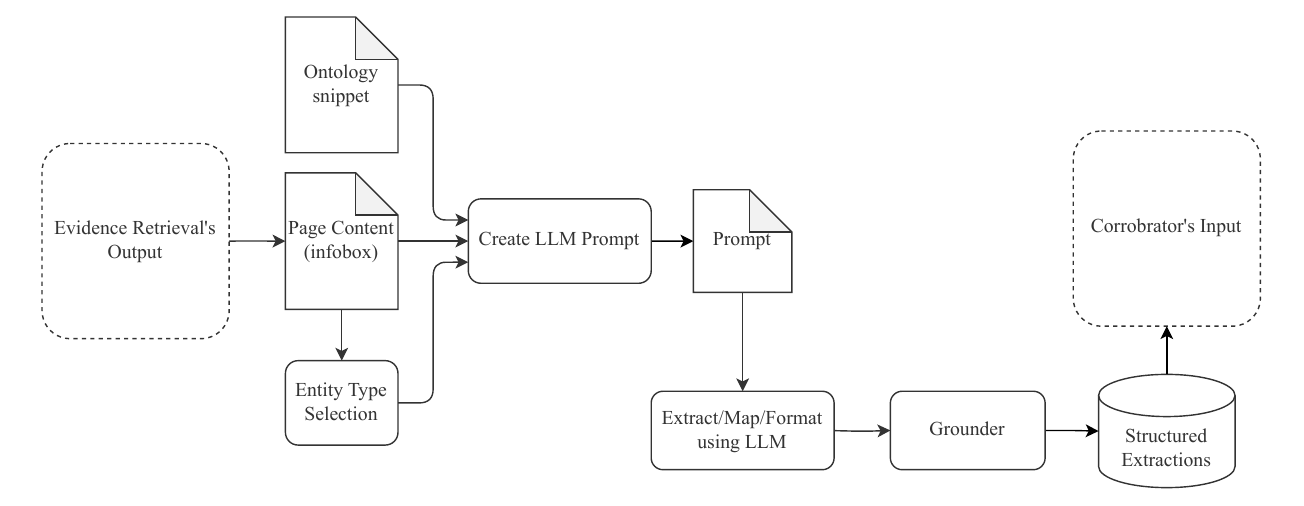}
    \vspace{-3mm}
  \caption{LLM-based extractor and grounder}
  \label{llm_extractor}
  \end{subfigure}
\caption{ODKE system architecture with main components including evidence retrieval, extraction, corroboration, and KG ingestion(a), and its LLM-based extractor(b)} \label{odke}
\end{figure}

The architecture of \system is illustrated in Figure~\ref{fig:architecture}. The system operates as a modular pipeline that begins with the \textit{Extraction Initiator}, which identifies missing or outdated facts in the knowledge graph or selects web sources for fact extraction. Next, the \textit{Evidence Retriever} gathers relevant documents from trusted sources such as Wikipedia.

\system then applies a series of extractors, pattern-based and LLM-based, to extract candidate facts. These extractions are passed to the \textit{Corroborator}, which normalizes, ranks, and consolidates the extracted values to produce high-confidence updates. 

While the system supports multiple extraction strategies, this paper focuses primarily on the \textbf{LLM-based knowledge extractor} (see Figure~\ref{llm_extractor}), which allows for flexible and scalable fact extraction across entity types.

\noindent
\textbf{Example.} Suppose the entity \textit{Taylor Swift}'s pages has been recently updated. The \texttt{Initiator} detects recent changes on her Wikipedia page. The \texttt{Retriever} fetches updated content. The LLM-based Extractor identifies "5 ft 11 in" for it's hight, maps it to the \texttt{height} predicate using ontology, and the Grounder validates that it is directly supported in the context. The Corroborator selects the most trustworthy variant and ingestion into the KG.

\noindent
In the following sections, we describe each component in more detail.
\subsection{Extraction Initiator}
\label{sec:initiator}
The \textit{Extraction Initiator} identifies candidate facts that are either missing or potentially outdated in the knowledge graph. It determines which entities require updates, and outputs them in a structured format for downstream processing. To identify stale or missing facts, \system leverages web activity signals. When a web page (e.g., a Wikipedia article) is updated or newly added, we treat this as a potential indication that associated entities may have new or changed information. Based on the subject(s) linked to the page, we schedule those entities for fact re-evaluation. This monitoring strategy enables timely detection of missing or outdated knowledge without exhaustive reprocessing of all KG subjects.

This module is designed to be flexible in its source selection. In our current setup, we prioritize Wikipedia pages that have been recently modified-based on the assumption that edits often reflect newly available or updated facts not yet present in the KG. In a time-lag analysis between Wikipedia and Wikidata, we found that facts on Wikipedia appear, on average, 69 days earlier. Commonly stale properties include \textit{height}, \textit{population}, \textit{age}, \textit{net worth}, \textit{weight}, \textit{unmarried partner}, \textit{child}, \textit{inception}, and \textit{date of birth}.

The output of this component is a database of tuples in the form \triple{$\langle$subject, URL, locale$\rangle$}. For instance, a sample tuple like \triple{$\langle$Michael Jordan, \href{https://en.wikipedia.org/wiki/Michael\_Jordan}{en.wikipedia.org/wiki/Michael\_Jordan}, En-US$\rangle$} instructs the system to extract relevant facts about Michael Jordan from the specified Wikipedia page.

\subsection{Evidence Retriever}
\label{sec:retriever}
Once subjects and source URLs have been selected by the Extraction Initiator, the \textit{Evidence Retriever} fetches web documents that are requested. In our current implementation, we operate under a controlled domain assumption, primarily using Wikipedia, where the mapping from entities to URLs is deterministic and language-specific.

For example, the entity \textit{Barack Obama} can be reliably mapped to its canonical Wikipedia URLs across multiple locales. Given this setup, evidence retrieval is reduced to fetching the relevant pages from a web crawl index, which serves as the document corpus for downstream extraction.

\subsection{Knowledge Extractor}
\label{arch:extractor}
Given a retrieved document and a target entity, the \textit{Knowledge Extractor} identifies and extracts the correct fact values (e.g., a date for \predicate{Date-of-Birth}, a currency amount for \predicate{Networth}, or a location for \predicate{Place-of-Birth}), and links them to their provenance spans in the source text.

\system employs two primary categories of extractors: \textbf{pattern-based} and \textbf{LLM-based}. Pattern-based extractors are optimized for high-precision, semi-structured data such as Wikipedia infoboxes, while LLM-based extractors generalize to a wider variety of contexts and data formats.

\paragraph{Pattern-based Extractors}
These extractors target semi-structured sources like Wikipedia infoboxes, and are designed to extract simple $(\textit{subject}, \textit{relation}, \textit{object})$ triples. Each infobox is interpreted as a set of key-value rows, where each row potentially represents a factual statement.

Pattern-based extraction uses manually defined rules, typically consisting of:
\begin{itemize}
    \item \textbf{Predicate Mapper:} Maps infobox keys to KG predicates. For example, the key \texttt{Height} may map to Wikidata property \texttt{P2048}.
    \item \textbf{Value Extractor:} Uses regular expressions to extract relevant values, supporting multiple formats (e.g., “6 ft 0 in” and “184 cm”) and hyperlinks to other entities.
    \item \textbf{Value Aggregator:} Consolidates multiple extracted candidates into a single, reliable value-filtering outliers or inconsistencies when necessary.
\end{itemize}

To ensure quality and consistency, each rule is validated against KG type constraints-for instance, ensuring that the object of property \texttt{P19} (Place-of-Birth) is an entity classified as a geographic location (e.g., \texttt{Q2221906} in Wikidata).

\paragraph{LLM-based Extractors}
The LLM-based extractor in \system is designed for broad, schema-aligned fact extraction from unstructured or semi-structured sources. Unlike pattern-based methods that depend on predefined fields, the LLM-based approach can generalize across entity types and attributes, making it highly adaptable for open-domain scenarios.

This component performs four core functions:
\begin{itemize}
\item \textbf{Prompt Generator:} Given a subject entity and its associated type(s), this module selects a relevant subset of the ontology-specifically, predicates applicable to that type-and converts it into a textual prompt. This ontology snippet includes predicate names and descriptions, required qualifiers(if any), and a set of acceptable units(if applicable), guiding the LLM to produce structured, semantically-aligned outputs. An example of the ontology-guided LLM prompt is provided in Appendix~\ref{sec:appendix-prompt}.
    
\item \textbf{Extraction:} The LLM uses the generated prompt and input context (e.g., Wikipedia content) to extract key-value pairs. It can also decompose composite values (e.g., extracting both "date" and "location" from a birth event).

\item \textbf{Schema Mapping:} The extracted values are aligned with appropriate predicates from the ontology snippet, using semantic cues from ontology.

\item \textbf{Schema Translation:} Finally, the structured extractions are serialized into a format suitable for downstream normalization and ingestion into the KG.
\end{itemize}

To support prompt generation in the LLM-based extractor, \system includes an offline pipeline that generates ontology-guided prompt templates-known as \textit{ontology snippets}-for each entity type. These snippets are used to guide LLM outputs toward semantically valid and schema-aligned facts.

\begin{figure}[ht]
    \centering
    \includegraphics[width=\linewidth]{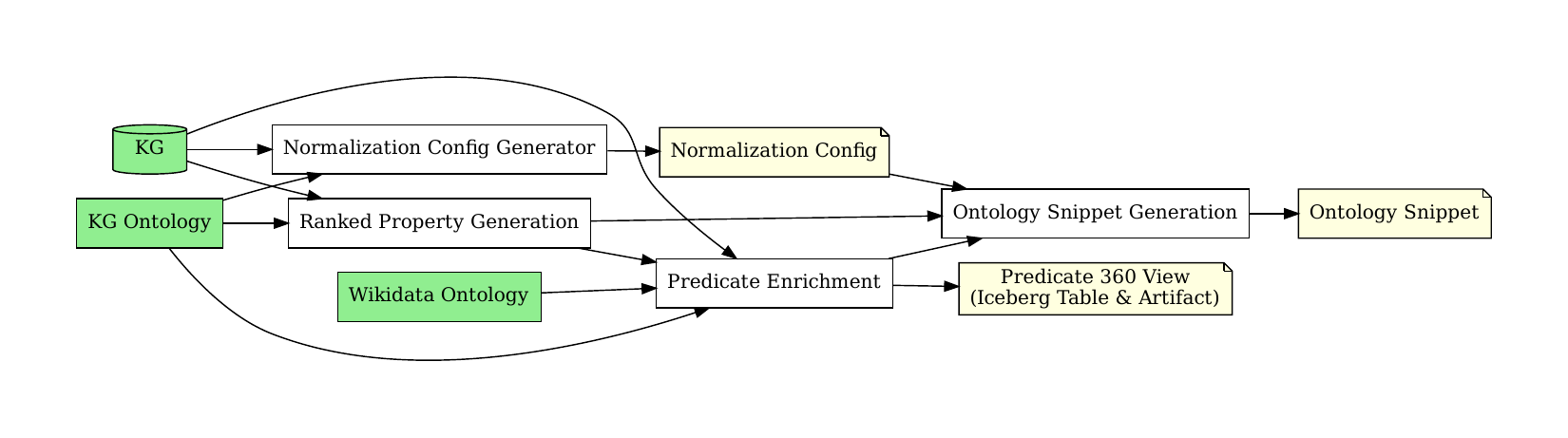}
    \caption{Ontology Snippet Generation pipeline. The system takes the KG, ontology, and Wikidata ontology as inputs and processes them through ranked property generation, predicate enrichment, and normalization configuration to produce a tailored ontology snippet for LLM prompting.}
    \label{fig:ontology-snippet-pipeline}
\end{figure}

The snippet generation pipeline as illustrated in ~\ref{fig:ontology-snippet-pipeline} includes two main components:

\begin{itemize}
    \item \textbf{Ranked Property Generation:} Given the KG, Ontology, this component computes a ranked list of predicates for each entity type. 
    To determine which properties(predicates) are supported for extraction, \system analyzes the ontology and computes a ranked list of important predicates per entity type. This ranking reflects the frequency and semantic importance of predicates observed in the KG and ensures that extraction is focused on the most salient and commonly populated attributes for each entity class.
    \item \textbf{Predicate Enrichment:} This component enriches ontology predicates by integrating information from multiple sources including ontology metadata for each predicate, Wikidata, and the KG. It produces an enriched view of predicates-termed \textit{Predicate 360}, as an artifact that includes:
    \begin{itemize}
        \item Predicate names from ontology
        \item Canonical labels, enriched descriptions
        \item Curated summaries (e.g., preferred value types, deterministic mappings)
        \item Domain, range, and maximum cardinality
        \item Associated qualifiers, ranked and pruned based on KG usage patterns
    \end{itemize}
\end{itemize}

The pipeline also includes a \textbf{Normalization Config Generator}, which identifies predicates requiring normalization (e.g., units for quantities like area, or formats for values like dates). It determines the list of expected units (e.g., square kilometers), formatting rules, and even provides value examples retrieved from the KG.

The output of this pipeline is an \textit{ontology snippet}-a curated, textualized schema fragment used as part of the LLM prompt. These snippets are refreshed periodically and optimized for token efficiency, enabling the LLM to perform schema-aware fact extraction that aligns with the KG structure and constraints and supports scalability across diverse entity types and adapts flexibly to both structured and unstructured web content.

Such a LLM-based extractor provides notable advantages, including scalability across all entity types and predicates without requiring customized configurations, as well as adaptability across multiple data modalities. 
\subsubsection{Grounder}
\label{arch:grounder}

The \textit{Grounder} is a filtering module that evaluates whether the facts extracted by the LLM-based extractor are truly supported by the evidence context provided. This step is crucial for ensuring factual accuracy and minimizing hallucinations that can arise from overgeneralized or inferred responses by large language models.

For each extracted triple, the Grounder constructs a natural language assertion and queries the grounder LLM to assess whether the assertion is grounded in the provided context. 

The judgment is made in a simple, interpretable format, typically:
\begin{itemize}
    \item \texttt{Yes}: the fact is explicitly supported by the context
    \item \texttt{No}: the fact is not supported\/cannot be verified
\end{itemize}

 Only those facts that receive an affirmative grounding judgment are retained for downstream corroboration. An example of the ontology-guided LLM prompt is provided in Appendix~\ref{sec:appendix-prompt_grounder}.

This approach ensures that the extracted knowledge remains verifiable and contextually anchored, adding a layer of factual control without introducing significant latency or computational cost.

\subsection{Corroborator}
\label{arch:corr}
After extraction and grounding, candidate facts are passed to the \textit{Corroborator}, which performs normalization, consolidation, and scoring to determine the most trustworthy values for each predicate to maximize factual precision while resolving potential conflicts across extraction strategies.

\system for a subset of extractions that still have to be normalized into their standardized forms according to their types, has a fallback strategy. For example, it uses Duckling \cite{Duckling} to convert expressions like “6 ft” or “July 4, 1776” into standardized numeric or temporal representations. Predicate-specific normalization logic is also applied when configured. Next, the Corroborator aggregates facts by predicate and subject, consolidating equivalent values and eliminating inconsistencies. It evaluates each candidate using a combination of heuristic and model-based scoring. Key features include:
\begin{itemize}
    \item Extractor type (pattern-based, LLM-based)
    \item Confidence or score from the extractor
    \item Frequency of the extracted value across evidence sources
    \item Richness of the extracted answer; such as inclusion of supporting qualifiers for the main fact
\end{itemize}

For predicates with low ambiguity, a simple rule-based ranking may suffice. For more complex scenarios, \system supports AutoML-based scoring using packages such as H2O \cite{H2OAutoML20} to learn weighting functions and rank facts accordingly.

\subsection{Data Export \& KG Ingestion}\label{subsec:ingestion}

The final stage in the \system pipeline involves ingesting validated and ranked facts into the knowledge graph. The exported output consists of high-confidence, normalized $\langle \text{subject}, \text{predicate}, \text{object} \rangle$ triples, along with optional qualifiers and metadata such as provenance and confidence scores.

These exported triples are treated as a new \textit{source} of data for the KG and are processed through the source ingestion pipeline, where they are aligned with the canonical ontology. 

When facts originate from sources with globally unique identifiers, such as Wikidata QIDs or Wikipedia URLs, the subject and object entities are directly linked to existing KG nodes. In contrast, facts extracted from unstructured web sources often lack such identifiers. In these cases, \system uses machine learning-based entity linking models to determine whether an existing entity in the KG matches the extracted one.

If no match is found, the system creates a new entity and inserts it into the KG, preserving the triple and linking it with minimal assumptions. This hybrid linking strategy ensures that the KG remains both semantically accurate and incrementally extensible.
\subsection{Deployment and Continuous Update}
\label{subsec:deployment}
\system supports two ingestion modes; while strict schema validation ensures consistency with the ontology:
\begin{itemize}
    \item \textbf{Batch mode}: periodic large-scale updates.
    \item \textbf{Streaming mode}: real-time ingestion of ongoing updates.
\end{itemize}

\subsection{Monitoring, QA, and Feedback Loops}
\begin{itemize}
    \item \textbf{Human curation (KG Quality team):}  
          A dedicated team audits $\approx$2\,000 randomly
          sampled triples each week.  \system must sustain
          $\geq 95\%$ precision to remain in production and has met
          this bar consistently since launch.
    \item \textbf{Side-by-Side (SBS) evaluation:}  
          Each week a stratified sample of real user KG queries is
          shown to annotators(shown anonymously to prevent bias) with the incumbent answer and the \system answer
          presented side-by-side. Annotators prefer the \system answer
          in roughly two-thirds of cases.
\end{itemize}
\section{Results and Impact}
\label{sec:results}
\system has been deployed in a production knowledge graph infrastructure since \textbf{May 2025}, running both batch and near-real-time streaming updates.  The pipeline is orchestrated via Kubernetes-managed Spark jobs triggered by Airflow DAGs. 
In production, \system processes an average of 150-250\,K new facts per day in streaming mode for high-priority entities, with end-to-end latency under 2 hours for most updates.

\subsection{Quality Evaluation and Component Impact}
\begin{itemize}
    \item \textbf{Precision:} Weekly audits consistently show over 95\% factual accuracy.
    \item \textbf{Coverage:} \system provides 10\% more answerable facts for user queries than prior pipelines.
    \item \textbf{Freshness:} Newly extracted facts appear roughly 50 days earlier than in legacy KG workflows.
    \item \textbf{Extensibility \& Reliability:} New predicates are integrated declaratively; schema validation and retry mechanisms keep success rates above 99.9\%.
    \item \textbf{Verification \& Ranking Impact:} Grounding reduced hallucinated extractions by 35\%. Corroboration improved factual precision from 91\% (raw LLM) to 98.8\% post-ranking.
\end{itemize}

\subsection{Production Metrics}
Table~\ref{tab:odke-metrics} summarizes key performance indicators from the latest deployment snapshot, demonstrating \system’s ability to scale reliably across diverse predicates and entity types while maintaining high factual precision.
\begin{table}[ht]
\centering
\begin{tabular}{|l|c|}
\hline
\textbf{Metric} & \textbf{Value} \\
\hline
Total facts extracted & 19M \\
Predicates supported & 195 \\
Pages with extractions & 4M \\
ODKE coverage vs. 3rd-party KG & 48\% \\
Precision & 98.8\% \\
\hline
\end{tabular}
\caption{\system Production Deployment Metrics}
\label{tab:odke-metrics}
\end{table}
\section{Conclusion}
\label{sec:conclusion}
We presented \system, a production grade system for open domain knowledge extraction using large language models. Building upon prior work, \system significantly extends the framework's capabilities with ontology-guided prompting, ontology oriented extraction in scale. The system also introduces a lightweight grounding verifier to ensure factual precision.

\system is currently deployed in a large-scale knowledge graph infrastructure, extracting and ingesting millions of facts across 195 predicates. Our results show strong precision and improved coverage relative to third-party knowledge sources, with support for both batch and real-time use cases.

We believe ODKE+ demonstrates the viability of LLM-based extraction in real-world KG pipelines and serves as a blueprint for deploying trustworthy, ontology-aligned fact extraction at scale.

\section*{Ethics Statement}
This work presents \system, a system for large-scale knowledge extraction using ontology-guided large language models. We recognize several ethical considerations associated with the development and deployment of such systems:
\begin{itemize}
\item \textbf{Bias Propagation:} Since \system relies on outputs from LLMs trained on broad web data, there is a risk of propagating social or cultural biases present in the training data. While our framework includes grounding and corroboration modules to ensure factual accuracy, mitigating underlying model biases remains a challenge and is an area for future work.
\item \textbf{Data Privacy:} ODKE+ exclusively processes public sources, such as Wikipedia, and is designed to extract only explicitly stated facts. It does not operate on user-generated content or private data, and its architecture avoids suppositional inference or hallucinated facts.
\item \textbf{Verification and Trustworthiness:} To mitigate hallucination risks, \system incorporates dedicated verification steps, including a grounding module that uses a second LLM to judge context alignment, and a corroborator that filters and scores extractions. These design choices have demonstrably improved factual precision in production.
\item \textbf{Environmental Impact:} While LLM-based systems can be computationally intensive, \system is optimized for hybrid operation, using lightweight pattern-based extractors where possible, and LLMs selectively, based on data complexity. Batch processing and pipeline orchestration are also optimized for cost and efficiency.
\end{itemize}

\section*{Acknowledgments}
We thank Mina Farid, Ameya Panse, Nandhitha Raguram, Mandana Saebi, and Pragnya Sridhar for their significant contributions to the software development underlying this work. Their efforts were essential in enabling our experiments and analyses.

\bibliography{custom}
\bibliographystyle{acl_natbib}

\appendix
\section{Sample Prompt for LLM-based Extraction}
\label{sec:appendix-prompt}

\noindent
Below is an example of an ontology-guided prompt used for extracting facts from a Wikipedia page. This prompt is fed to the LLM along with an entity-specific ontology snippet and web page content.

\begin{quote}
\scriptsize
You are an accurate information extraction system. Your task is to extract and map facts from a provided input passage. The mapping will be guided by a provided input ontology.

\textbf{Response Format:}
\begin{itemize}
  \item Return the extracted facts in a valid JSON dictionary.
  \item Keys must match properties from the Input Ontology with supporting mentions in the passage.
  \item Values should be lists of JSON objects. Each object may include:
  \begin{itemize}
    \item \texttt{"answer"} (required): Shortest surface form from passage.
    \item \texttt{"qualifiers"} (optional): Dictionary of related qualifier fields.
    \item \texttt{"normalized answer"} (optional): If normalization is required per ontology.
    \item \texttt{"normalization unit"} (optional): Corresponds to normalized value unit.
  \end{itemize}
\end{itemize}
\textbf{Important Notes:}
\begin{itemize}
  \item Extract only facts explicitly mentioned in the passage.
  \item Do not infer or hallucinate facts or retrieve from memory.
  \item Validate and return only syntactically correct JSON.
\end{itemize}
\textbf{\# Start Input Passage}
\begin{verbatim}
Title: Felton Ross
Infobox properties: Dr. Felton Ross:

{"Born": "William Felton Ross\nMay 9, 1927\nLeominster, England"}
{"Died": "March 3, 2022 (aged 94)\nGoshen, Indiana"}
{"Nationality": "British-American"}
{"Education": "London Hospital Medical College (1954)"}
{"Occupation(s)": "Physician, christian missionary"}
{"Spouse": "Una Dickinson (m. 1959)"}
{"Children": "5"}
{"Medical career\nResearch": "Leprosy research"}
\end{verbatim}
\textbf{\# End Input Passage}
\noindent
\textbf{\# Start Input Ontology} \\
\begin{verbatim}
date of birth: Property relating an organism to its date of birth.
place of birth: Property relating an organism to its place of birth.
date of death: Property relating an organism to its date of death.
occupation: Property relating a person to one of their occupations.
educated at: Property relating a person to their place of education., 
    qualifiers: [academic degree, start time, end time]
spouse: Property relating a person to their marriage relationships., 
    qualifiers: [start time, end time, end cause]
child: Property relating a parent to their child.
height: Property relating an entity to its height, 
    need\_normalization: True, normalization\_unit: centimetre
...
\end{verbatim}
\textbf{\# End Input Ontology}
\end{quote}

\section{Sample Prompt for LLM as a judge (Grounder)}
\label{sec:appendix-prompt_grounder}

\noindent
Below is an example of a prompt used for verifying the extracted facts from a Wikipedia page are grounded in the context. This prompt is fed to the second LLM(\textit{grounder}).

\begin{quote}
\scriptsize
 Given a context about a subject and a triple in the format of <subject, predicate(qualifier: optional), object>, your task is to verify if the given triple can be found from or grounded in the context and only respond with True or False.
For some object, they may have been listed there in form of a list of object, not single.

\textbf{**Context:}
\begin{verbatim}
Title: Felton Ross
Infobox properties: Dr. Felton Ross:

{"Born": "William Felton Ross\nMay 9, 1927\nLeominster, England"}
{"Died": "March 3, 2022 (aged 94)\nGoshen, Indiana"}
{"Nationality": "British-American"}
{"Education": "London Hospital Medical College (1954)"}
{"Occupation(s)": "Physician, christian missionary"}
{"Spouse": "Una Dickinson (m. 1959)"}
{"Children": "5"}
{"Medical career\nResearch": "Leprosy research"}
\end{verbatim}
\textbf{**triple:} 
\begin{verbatim}
<Felton Ross, Date of Birth, May 9, 1927>
\end{verbatim}
\end{quote}

\end{document}